# Humans and Large Language Models in Clinical Decision Support: A Study with Medical Calculators

**Nicholas C. Wan, BE[1], Qiao Jin, MD[1], Joey Chan, BS[1], Guangzhi Xiong, BA[2], Serina Applebaum, BS[3], Aidan Gilson, MD[4], Reid McMurry, MD[5], R. Andrew Taylor, MD, MHS[6], Aidong Zhang, PhD[2], Qingyu Chen, PhD[7], Zhiyong Lu, PhD[1]**

[1]Division of Intramural Research, National Library of Medicine (NLM), National Institutes of Health (NIH), Bethesda, MD, USA; [2]Department of Computer Science, University of Virginia, Charlottesville, VA, USA; [3]Department of Ophthalmology & Visual Science, Yale School of Medicine, New Haven, CT, USA; [4]Massachusetts Eye and Ear, Harvard Medical School, Boston, MA, USA; [5]Department of Emergency Medicine, Boston Medical Center, Boston, MA, USA; [6]Department of Emergency Medicine, University of Virginia, Charlottesville, VA, USA; [7]Department of Biomedical Informatics & Data Science, Yale School of Medicine, New Haven, CT, USA

**Abstract**

*Although large language models (LLMs) have been assessed for general medical knowledge using licensing exams, their ability to support clinical decision-making, such as selecting medical calculators, remains uncertain. We assessed nine LLMs, including open-source, proprietary, and domain-specific models, with 1,009 multiple-choice question-answer pairs across 35 clinical calculators and compared LLMs to humans on a subset of questions. While the highest-performing LLM, OpenAI's o1, provided an answer accuracy of 66.0% (CI: 56.7-75.3%) on the subset of 100 questions, two human annotators nominally outperformed LLMs with an average answer accuracy of 79.5% (CI: 73.5-85.0%). Ultimately, we evaluated medical trainees and LLMs in recommending medical calculators across clinical scenarios like risk stratification and diagnosis. With error analysis showing that the highest-performing LLMs continue to make mistakes in comprehension (49.3% of errors) and calculator knowledge (7.1% of errors), our findings highlight that LLMs are not superior to humans in calculator recommendation.*

**Introduction**

Large language models (LLMs) such as GPT-4 and Med-PaLM have achieved expert-level performance on biomedical tasks like medical question-answering (QA)[1–5]. While these LLMs have been evaluated using medical licensing exams to quantify their general medical knowledge[6–8], their capability to provide real-world clinical decision support (CDS) and to go beyond factual recall cannot be fully assessed by medical exams alone[9]. Simultaneously, clinical calculators, i.e. evidence-based computational tools for risk assessment, prognosis, and diagnosis, play an important role in real-world clinical care delivery[10,11] and are becoming increasingly prevalent via web-based platforms like MDCalc[12,13]. With recent research also showing that LLM agents can engage in tool learning[14,15], clinical calculators represent a set of tools that may augment medical LLM agents and their capability to probabilistically reason[16]. However, the capability of LLMs to reason through complex clinical scenarios and appropriately select calculators for recommendation remains understudied. Thus, our study aims were (1) to explore whether LLMs can accurately recommend calculators in a variety of clinical scenarios and (2) to compare LLM performance with medical trainees. We hypothesized that LLMs would demonstrate comparable or superior accuracy to medical trainees when recommending relevant clinical calculators via our benchmarking dataset.

To address these study aims, we conducted the first comprehensive exploratory analysis of whether LLMs can recommend clinical calculators, with comparison to human performance, and curated MedQA-Calc, a first-of-its-kind dataset for evaluating the clinical calculator recommendation capabilities of LLMs. To construct MedQA-Calc, we identified 35 popular clinical calculators from MDCalc.com, excluding questionnaire calculators as they can rely on subjective inputs not found in patient notes. We curated questions relating to these calculators using publicly available case reports in PubMed Central (PMC-Patients)[17]. We identified instances of calculator usage within patient cases, removed evidence of respective calculator usage, and constructed multi-choice questions using extracted calculators as ground-truth. We curated 1,009 question-answer instances, where each instance contains a truncated patient case, a generalized recommendation question, answer choices, and a ground-truth answer.

**Methods**

*Data Sources and Processing:* Figure 1 presents the overall study design. Our data source, PMC-Patients (derived from PubMed Central), covers a broad range of clinical settings (emergency medicine, surgery, and inpatient/outpatient care) and includes patient cases from the United States and international populations. Using this source for anonymized information, we leveraged GPT-4o and few-shot learning to extract clinical calculators from patient cases (code at https://github.com/ncbi-nlp/Calculator-Recommendation). We then truncated extracted texts to remove evidence of calculator usage via a fine-tuned model of GPT-4o. Next, we curated questions by incorporating five answer choices from our list of calculators, excluding calculator options that were similar to the ground-truth calculator. To assess our benchmark, we focused on three settings: (1) *LLM performance*, where models answered the entire question set; (2) *Human performance*, where two medical trainees answered a subset of questions; and (3) *Error analysis*, where we reviewed errors made by LLMs on questions used for human evaluation.

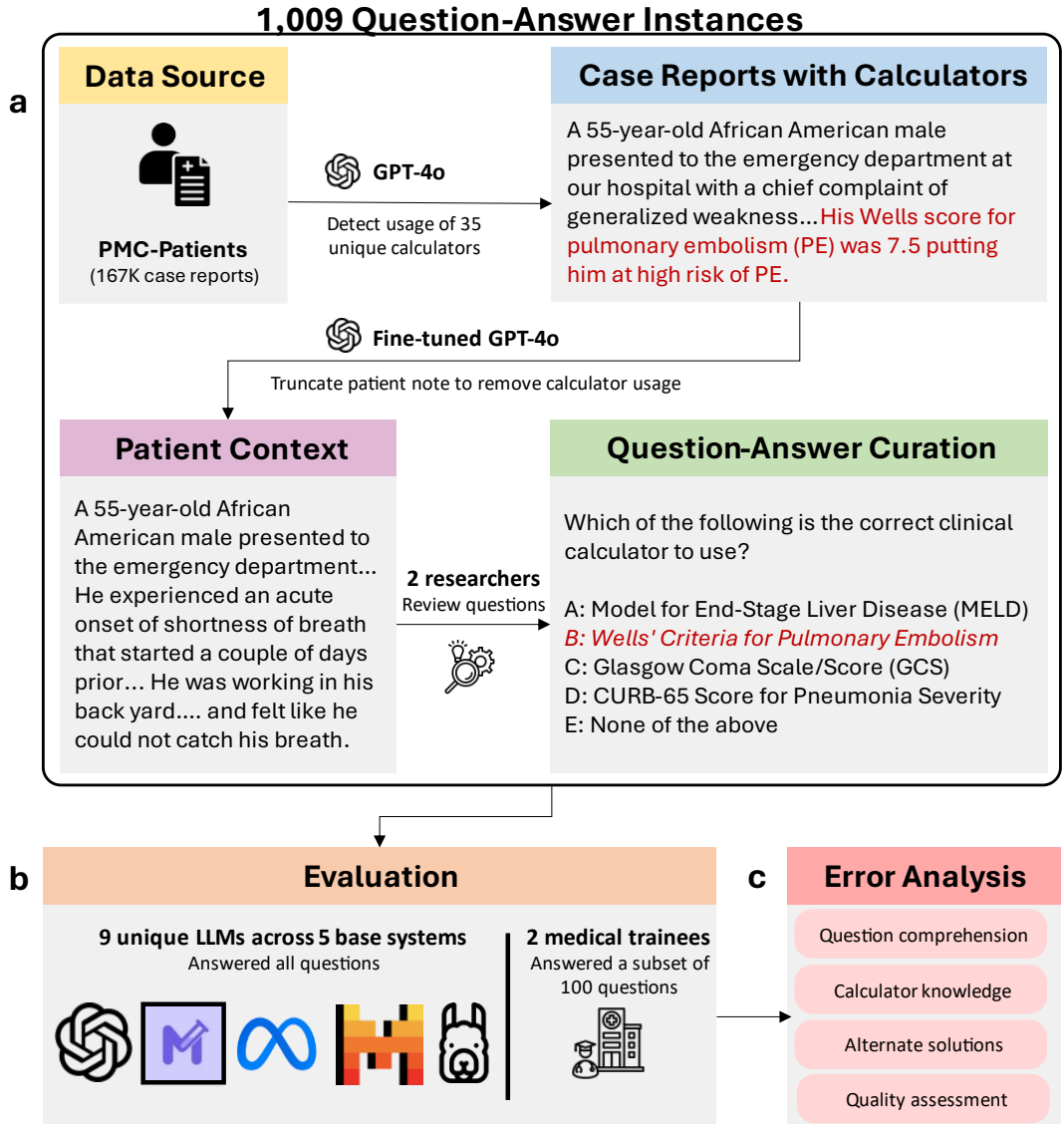

**Figure 1.** Study design. The curation and evaluation workflow for our medical calculator question-answering dataset, MedQA-Calc. **a.** Patient case reports containing calculators were identified with GPT-4o and transformed into questions using a fine-tuned GPT-4o model. Answer options were curated using automatic randomization, excluding similar answer choices. **b.** The capability of LLMs and medical trainees to answer these questions and recommend calculators was evaluated. **c.** Researchers and medical trainees reviewed LLM errors and evaluated the quality of questions.

*Calculator Selection:* We manually reviewed MDCalc.com in June 2024 to identify 35 medical calculators (Table 1) from MDCalc's internal "Popular" list of calculators, to collect calculator names, and to curate calculator codes (e.g., abbreviating "Glasgow Coma Scale" as "gcs"). While popular calculators involving only questionnaire data were excluded from the evaluation, we gathered calculators across medical domains including cardiology, neurology, pulmonology, critical care, and surgery. Next, we extracted patient summary information from PMC-Patients, a publicly available dataset of real, anonymized case reports.

**Table 1.** Calculator examples. Ten selected calculators from the 35 extracted medical calculators leveraged in this study, including abbreviations found in Figure 2 and their codes used with LLMs.

| Calculator Name | Calculator Abbreviation (Code) |
| --- | --- |
| The Acute Physiology and Chronic Health Evaluation II (APACHE II) score | APACHE II score (apache) |
| CHA2DS2-VASc Score for AF | CHA2DS2-VASc Score for AF (cha2ds2) |
| Child-Pugh Score for Cirrhosis Mortality | Child-Pugh Score for Cirrhosis Mortality (child_pugh_score) |
| CURB-65 Score for Pneumonia Severity | CURB-65 Score (curb65) |
| Fibrosis-4 (FIB-4) Index for Liver Fibrosis | Fibrosis-4 Index for Liver Fibrosis (fib4) |
| HEART Score | HEART Score (heart) |
| Model for End-Stage Liver Disease (MELD) Score | MELD Score (meld) |
| National Institutes of Health Stroke Scale/Score (NIHSS) | NIHSS (nihss) |
| Sequential Organ Failure Assessment (SOFA) Score | SOFA Score (sofa) |
| Wells' Criteria for Pulmonary Embolism | Wells' Criteria for Pulmonary Embolism (wells_pe) |

*Note Extraction:* Using GPT-4o, we automatically extracted text evidence of calculators within each patient note, patient ID, calculator score values, calculator names, and calculator units. After data cleaning (i.e., limiting to 1000 instances per calculator, removing calculators not found in our list of 35 calculators), our dataset, MedQA-Calc, consisted of 7,768 medical calculator prediction instances.

*Note Truncation:* Then, we implemented a GPT-4o model (fine-tuned on 70 patient-calculator note instances manually truncated by NW) to truncate patient notes at the appearance of any given extracted medical calculator name and value. An example of note truncation is shown below in Table 2.

**Table 2.** Patient vignette processing. Text is truncated such that evidence of a given medical calculator is removed.

| Note Type | Text |
| --- | --- |
| Original | A 55-year-old African American male with a past medical history of hypertension presented to the emergency department at our hospital with a chief complaint of generalized weakness… His Wells score for pulmonary embolism (PE) was 7.5 putting him at high risk of PE. |
| Truncated | A 55-year-old African American male with a past medical history of hypertension presented to the emergency department at our hospital with a chief complaint of generalized weakness… |

*Question Curation:* After running notes through the truncation model, we had a total of 7,758 medical calculator question-answer instances. We then randomly split this data into a training and test set of 6,614 and 1,144 instances, respectively. We further curated the testing set by filtering across all calculator types, ensuring a maximum 50 instances per calculator. Next, two researchers, NW and JC, manually reviewed the test instances to ensure that notes were properly truncated with respect to the original note and calculator evidence. After NW reviewed all instances, JC reviewed all validated instances. After manual review, question-answer vignettes were automatically assigned answer options (answer choices) to include five options (A, B, C, D, or E). Of note, we included "None of the above" as a possible answer choice (E) in all cases. In 80% (4/5) of questions, the correct answer choice was randomized to A, B, C, or D, and incorrect labels were randomized to the unfilled choices. In 20% (1/5) of questions, we included "None of the above" as the correct answer choice, removing the ground truth answer choice as an option. We also created an exclusion matrix so that calculators with similar purposes (e.g., CURB-65 and Pneumonia Severity Index) were not provided alongside one another.

*Question Evaluation:* We evaluated nine LLMs (i.e., OpenAI o1 (version 2024-12-01), GPT-4o (version 2024-02-01), GPT-3.5-Turbo (version 2024-02-01), Llama-3-70B, Llama-3-8B, Mixtral-8x7B, Mistral-7B, Meditron, and PMC-LLaMA) using our dataset of 1,009 questions. We chose these nine LLMs to capture the performance of proprietary, open-source, and medical domain-specific models. Additionally, we employed two medical trainees (one medical student and one first-year medical resident) to review 100 instances of calculator recommendations across all 35 calculators. Medical trainees answered questions in a closed book setting with no external resources. After receiving ground-truth answers, medical trainees then evaluated calculator instances by labeling their agreement with the question-answer vignette.

*Error analysis:* To perform additional analysis of LLM errors, we had two annotators (one research trainee, NW, and one medical student, SA) review errors made by LLMs on the subset of 100 questions answered by the medical trainees. Annotators labeled LLM errors as (1) comprehension errors, (2) calculator knowledge errors, (3) alternate solution errors, or (4) "no explanation" errors according to our provided annotation guidelines. Of note, "no explanation" errors were defined as errors made by LLMs that did not provide an explanation or rationale for answers. In cases of error type disagreement, annotators discussed the LLM error until a consensus was reached.

*Statistical Testing:* To quantify the performance of different LLMs on our dataset of 1,009 questions, we performed McNemar's test for all pairwise comparisons between the nine different models; to adjust for multiple testing, we utilized the Bonferroni correction to calculate a significance threshold of 0.05 / 36 (the total number of test comparisons between different LLMs).

For human evaluation, we performed McNemar's test, using an alpha of 0.0167 (i.e., 0.05 / 3 test comparisons), to make pairwise comparisons between trainees and the highest performing LLM, OpenAI's o1, on a subset of 100 questions. We also calculated the 95% confidence intervals (CIs) associated with the accuracy of the medical trainees and o1. Additionally, we leveraged bootstrapping with 10,000 samples to calculate the average answer accuracy of the medical trainees and its associated 95% confidence interval.

**Results**

Figure 2 presents the distribution of calculators or selection options within the MedQA-Calc dataset. Common calculators and rule-based tools included the CHA2DS2-VASc Score for Atrial Fibrillation, APACHE II Score, and the National Institutes of Health Stroke Scale (NIHSS). Examples of less frequent tools include the Atherosclerotic Cardiovascular Disease (ASCVD) risk calculator, the Centor Score, and the Pulmonary Embolism Rule-out Criteria (PERC).

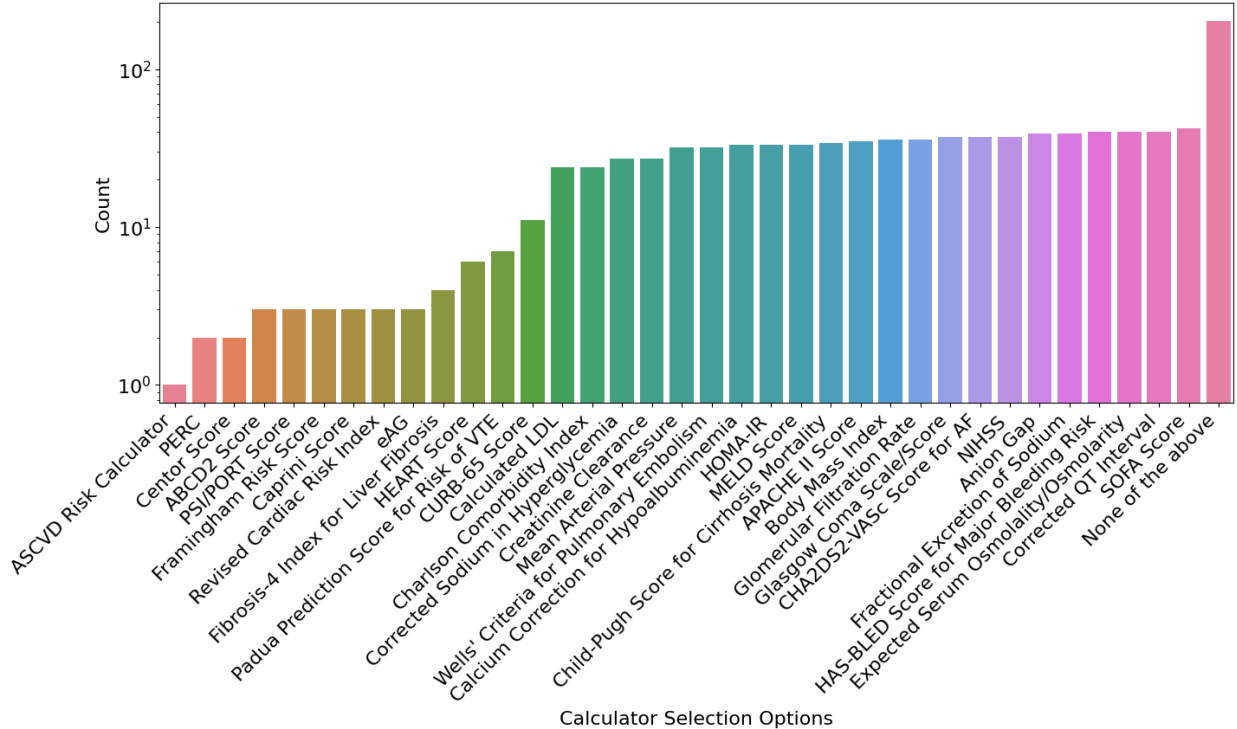

**Figure 2.** Extracted medical calculators. The frequency of clinical calculators found in the MedQA-Calc dataset.

The results of the evaluation are shown in Figure 3. LLM and human performance were evaluated in "closed book" settings as neither evaluation included the use of external tools like literature search engines or web browsing. Nine LLMs were leveraged for evaluation, and two medical trainees participated in human evaluation. Within LLM performance, o1 provided the best nominal stand-alone performance; however, o1 performance is not significantly different from Llama-3-70B or GPT-4o performance. Additionally, the low performance of PMC-LLaMA suggests that data contamination alone does not improve answer accuracy beyond the human baseline as PMC-LLaMA has been trained on the corpus containing PMC-Patients[18]. With human performance, two medical trainees provided the same answer choices for 69 out of 100 questions. One medical trainee (90.0%, CI: 84.1-95.9%) significantly outperformed o1 (66.0%, CI: 56.7-75.3%) on a subset of questions, and one medical trainee (69.0%, CI: 59.9-78.1%) matched performance with o1. After bootstrapping, the average answer accuracy of the medical trainees was 79.5% (CI: 73.5-85.0%). Given this variability in human performance, LLMs show the potential to provide calculator recommendations on par with humans; however, they have not surpassed human-level performance, even at the medical training level, for the task of clinical calculator selection. To investigate errors made by LLMs, we divided LLM errors into four groups: comprehension (misunderstanding or hallucination), calculator knowledge (incorrect calculator usage), alternative solution (choosing a different calculator option that may be appropriate), and no explanation (LLMs lacking sufficient explanation). To assess human error, medical trainees reviewed their errors and reported agreement with ground-truth answers. Reasons for disagreement included limited patient note vignettes and alternative relevant answer options.

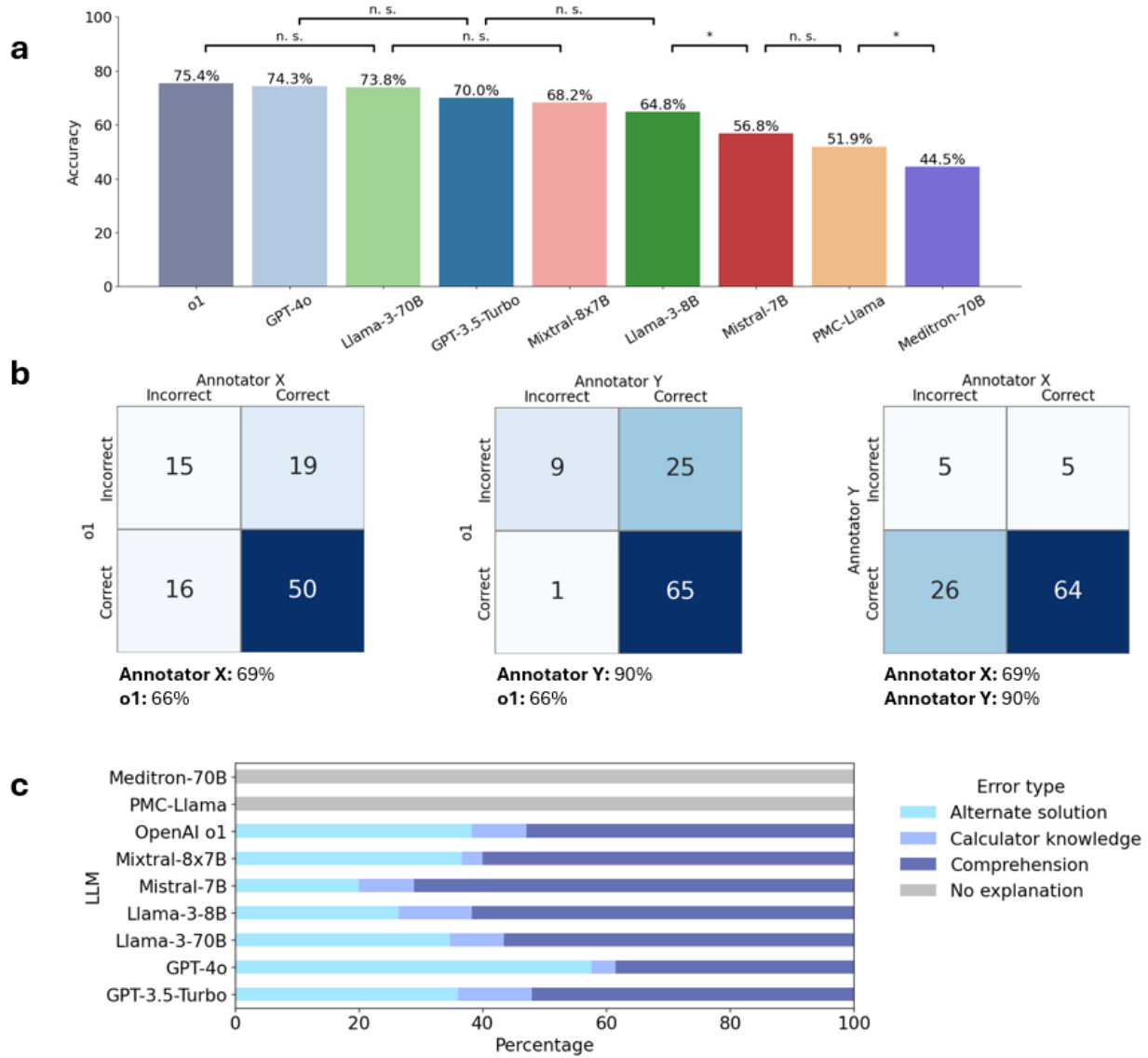

**Figure 3.** Evaluation results. Large language model performance, human performance, and error analysis. **a.** The accuracy of different LLMs in answering our curated questions; n.s. not significant, *, significant with p < 0.05 /36. pairwise accuracies are significantly different, if not indicated by n.s. labels. **b.** Confusion matrices representing the performance of two medical trainees and the highest-performing LLM on a subset of 100 questions. **c.** Bar graphs representing the proportions of errors made by each LLM model on a subset of 100 questions, stratified by error type (i.e., comprehension, calculator knowledge, alternative solution, no explanation).

Figure 3b displays the confusion matrices of o1 and medical trainees. Depending on annotator comparison, o1 correctly answered at least one out of the 100 questions that medical trainees incorrectly answered. We also found that high-performing models, like o1, GPT-4o, or Llama-3-70B, commonly made comprehension errors while also suggesting alternative answers that were deemed appropriate (Fig. 3c). In contrast, high-performing models made proportionally fewer errors due to incorrect calculator knowledge.

**Discussion and Conclusions**
*Benchmarking LLM and human performance:* Our analyses show performance trends that provide insight into the medical reasoning and calculator recommendation capability of large language models and humans. Among the studied large language models, for example, OpenAI's o1, GPT-4o, and Llama-3-70B provide the greatest accuracy when generating explanations and answers for calculator recommendation questions based on patient vignettes. At the same time, medical trainees, or human annotators, perform comparably or better on a subset of 100 questions, compared to the highest performing LLM, o1. Notably, o1 offered a nominally lower answer accuracy of 66.0% on the subset of questions as compared to its overall answer accuracy of 75.4%.

*Error Analysis:* A closer look at explanations during error analysis revealed that advanced models may struggle with making inferences with respect to time. For example, GPT-4o may automatically assume that a patient, who has already received a Sequential Organ Failure Assessment (SOFA) score, does not require an additional SOFA score. Moreover, GPT-4o may fail to recognize the utility of using a calculator score for diagnostic support. If a truncated patient note reveals a respiratory diagnosis, GPT-4o may report that a respiratory risk calculator is not needed as a diagnosis has already been made. Furthermore, the highest-performing model, o1, also revealed errors in calculator knowledge. For example, when posed with a patient eligible for the Atherosclerotic Cardiovascular Disease (ASCVD) Risk calculator, o1 incorrectly reported that the ASCVD calculator is only used for patients without ASCVD history; one possible explanation for this incorrect assumption is that multiple versions of the ASCVD risk calculator exist online (LLM explanations and data are publicly available at https://github.com/ncbi-nlp/Calculator-Recommendation). Ultimately, these example errors suggest that large language models may require further integration with time-series data and domain-specific knowledge to provide applicable, real-world clinical decision support.

*Related Work*: Existing works for benchmarking LLMs in medicine include PubMedQA, MedQA, MedMCQA, and medical questions found in MMLU[6-8,19]. Like these datasets, which evaluate the qualitative reasoning of LLMs, MedQA-Calc evaluates the capability of LLMs in selecting medically relevant solutions based on patient cases. In contrast to these datasets, the MedQA-Calc benchmark is solely focused on tool selection, and our work highlights that medical tool selection is often non-trivial, requiring domain expertise to select the most appropriate and beneficial tool.

Research primarily focused on both medical calculators and large language models includes AgentMD, MedCalc-Bench, and OpenMedCalc[16,20,21]. AgentMD, for example, focuses on automated clinical calculator curation and evaluation. Through the development of a framework using language agents as tool makers and tool users, AgentMD offers a promising approach for risk calculation and AI-assisted clinical decision-making; however, further research with AgentMD across diverse patient populations is required to assess its strengths and weaknesses. Additionally, evaluations from MedCalc-Bench, which measured LLM quantitative medical reasoning, highlight that LLMs are not yet computationally accurate or reliable enough for clinical use with prompt engineering alone. Finally, OpenMedCalc from Goodell et al. has developed an open-source clinical calculation application programming interface (API) integrated with ChatGPT, and more recently, work from Goodell et al. demonstrates that the medical calculation output of LLMs can be improved through the use of both generic and task-specific tools[22]. Our question set, MedQA-Calc, builds on these previous works by offering both preprocessed and manually reviewed natural language patient cases with corresponding evidence of calculator usage.

*Limitations:* Our study design has several limitations. Firstly, though we have manually evaluated the relevance of each calculator for each instance, clinical calculator selection may not always be necessary given evidence-based practices. In addition, though we designed our study to exclude similar calculators during answer choice curation, multiple distinct medical calculators may be appropriate in clinical practice. For example, the process of truncating patient reports at the first appearance of a given medical calculator can remove mentions of other pertinent calculators within downstream text. Moreover, the size and scope of the dataset is limited as it uses only 35 medical calculators; thus, we excluded other medical calculators that may have appeared within patient texts. Next, the PMC-Patients dataset is composed of case reports which may deviate from clinical scenarios where clinical calculators are more commonly applied; such case reports also span multiple countries, where disease burden, documentation, and practice patterns may vary. In terms of evaluation with large language models, we utilized a traditional baseline chain-of-thought prompting strategy (e.g., "generate your output as an explanation followed by an answer"), and we did not explore alternative prompting strategies such as retrieval augmented generation (RAG) or few-shot learning (FSL), where prompts are modified via search engine / API integration or via the use of exemplars, respectively[23-25]. Additionally, we did not use other advanced reasoning models such as OpenAI's o3-mini, DeepSeek-R1, or Gemini 2.0 Flash Thinking, which may yield substantially different results[26-28]. We also identified variability in human annotators, with human performance varying over 20% between two trainees. Given this variability, additional human evaluators may provide a more robust comparison. Finally, our dataset relies on the use of multi-choice answering. In real-world settings, providers are not given multiple options to select, and clinical workflow can require open-ended discussion and synthesis.

*Future work:* In future work, we may address these limitations by expanding the coverage of clinical calculators or medical decision-making scenarios and creating more granular patient vignettes, stratified across medical specialties. Additionally, we may consider curating open-ended clinical calculator questions to further evaluate the clinical utility of LLMs in medicine; more specifically, we may explore curating questions that have a well-defined calculator option that is more relevant than similar calculators or tools. In other words, we may integrate challenging negative cases within calculator selection vignettes to test a model's ability to choose the best possible tool. We may also measure the baseline quantitative reasoning of LLMs as done in recent work with MedCalc-Bench, a benchmarking study on the capability of LLMs to perform medical calculations[20]. Finally, we may further integrate our dataset with LLMs through tool learning, a key feature of language agents. With works like OpenMedCalc and AgentMD using medical calculators to augment LLMs[16,21], our dataset of explanations and calculator evidence may aid in the fine-tuning of language agents with medical calculators.

*Conclusions:* In summary, we present the curation and evaluation of question-answer instances relating to medical calculator usage, a complex clinical task. In addition to o1 demonstrating nominally lower multi-choice accuracy as compared to the average performance of medical trainees, large language models continue to show errors in comprehension and calculator knowledge. Despite filtering similar calculators during question-answer curation, our research also identified cases in which other calculator answers may be appropriate. This suggests that further curation and evaluations in open-ended settings are needed to test the capabilities of LLMs in medical decision-making.

**Author contributions**
Study concepts/study design, N.W., Q.J., Q.C., Z.L.; paper drafting or paper revision for important intellectual content, all authors; approval of the final version of the submitted paper, all authors; agrees to ensure any questions related to the work are appropriately resolved, all authors; literature research and review, N.W., Q.J., Z.L.; experimental studies, N.W., Q.J., G.X., A.Z..; human review or evaluation, N.W., J.C., S.A., A.G.; data interpretation and statistical analysis, N.W., Q.J., R.M., A.T., Q.C., Z.L.; and paper editing, all authors.

**Acknowledgements**
This research was supported by the Division of Intramural Research (DIR) of the National Library of Medicine (NLM), National Institutes of Health.

**Conflicts of interest**
The authors declare no competing interests.

**Data availability**
PMC-Patients is available via Hugging Face datasets at https://huggingface.co/datasets/zhengyun21/PMC-Patients.

The MedQA-Calc dataset is available at https://huggingface.co/datasets/Nicholas-Wan/MedQA-Calc.

Example code is available at https://github.com/ncbi-nlp/Calculator-Recommendation.